\documentclass{style/svproc}
\usepackage{url}

\usepackage{booktabs}
\usepackage{longtable}
\usepackage{graphicx}
\usepackage{lscape}
\usepackage{multirow}
\usepackage{enumitem}
\usepackage{url}
\usepackage{hyperref}

\begin{document}
\mainmatter              
\title{Taxonomy and Survey on Remote Human Input Systems for Driving Automation Systems}
\titlerunning{Taxonomy on Remote Human Input Systems}  
%
\author{Daniel Bogdoll \and Stefan Orf \and Lars Töttel \and J. Marius Zöllner}
%
\authorrunning{Daniel Bogdoll et al.} 
%
\tocauthor{Daniel Bogdoll, Stefan Orf, Lars Töttel, and J. Marius Zöllner}

%
\institute{FZI Research Center for Information Technology, Germany\\
\email{\{bogdoll, orf, toettel, zoellner\}@fzi.de}}


\maketitle              

\begin{abstract}
    Corner cases for driving automation systems can often be detected by the system itself and subsequently resolved by remote humans. There exists a wide variety of approaches on how remote humans can resolve such issues. Over multiple domains, no common taxonomy on those approaches has developed yet. As the scaling of automated driving systems continues to increase, a uniform taxonomy is desirable to improve communication within the scientific community, but also beyond to policymakers and the general public. In this paper, we provide a survey on recent terminologies and propose a taxonomy for \emph{remote human input systems}, classifying the different approaches based on their complexity.

\end{abstract}

\section{Introduction}
Since the scaling of automated driving systems\footnote{We follow the SAE J3016\textsubscript{202104} standard \cite{sae_j3016c_2021}, where \emph{driving automation system} (DAS) refers to level 1-5, \emph{automated driving system} (ADS) to level 3-5, and \emph{driver support} (DS) to level 1-2 systems. Details on the levels can be found in the standard.} (ADS) is progressing rapidly, with first level 4 systems publicly available \cite{ackerman2021WhatFullAutonomy}, dealing with rare situations such as corner cases is becoming more relevant. While the maturity of ADS is steadily increasing, such situations still need human support. Thus, many systems are capable of detecting situations in which they are unsure on how to proceed. In such situations they can receive input from a remotely located human, which we refer to as \emph{remote operator} (ROp). There are many parties discussing such \emph{remote human input systems} (RHIS) - as a result, a wide variety of different terms have been used within this domain. This paper offers a survey on existing literature, approaches and terminologies and provides the research community with a taxonomy on \emph{remote human input systems}.


\section{Survey on Remote Human Input Systems}
\label{sec:survey}

We classify existing approaches based on the SAE J3016\textsubscript{202104} levels \cite{sae_j3016c_2021}. While these are well known, it is important to highlight that a DAS-equipped vehicle does not necessarily perform on a constant level during any given trip. For instance, a DAS which is currently running at level 4, might encounter a situation it can not handle anymore. While the DAS would continue on level 4 with a successful \emph{remote assistance}, it would degenerate to the level range 1-2 if \emph{remote driving} would be necessary for recovery. An overview can be found in Table \ref{tab:survey}. The survey is divided into four categories: Standards and technical reports, corporate approaches, contributions from the scientific community and legislative bills. In the first column, the source including the utilized, general term for RHIS is shown. The second and third column follow the same pattern, while differing in the respective SAE level. In each cell, we first list the task-related terminology, followed by the human-related terminology.


\begin{table}[]
	\resizebox{\textwidth}{!}{%
\begin{tabular}{lcc}
\hline
\multicolumn{1}{|l|}{\textbf{Source}}                                                                                       & \multicolumn{1}{c|}{\textbf{Level 1-2}}                                                                                                       & \multicolumn{1}{c|}{\textbf{Level 4-5}}                                                                                      \\ \hline
\multicolumn{1}{|l|}{SAE \cite{sae_j3016c_2021}}                                                                                                   & \multicolumn{1}{c|}{Remote Driving}                                                                                                           & \multicolumn{1}{c|}{Remote Assistance}                                                                                       \\
\multicolumn{1}{|l|}{\textbf{Remote Support Functions}}                                                                     & \multicolumn{1}{c|}{Remote Driver}                                                                                                            & \multicolumn{1}{c|}{Remote Assistant}                                                                                        \\ \hline
\multicolumn{1}{|l|}{CSCRS \cite{cummings2020ConceptsOperationsAutonomous}}                                                                                                 & \multicolumn{1}{c|}{Teleoperation}                                                                                                            & \multicolumn{1}{c|}{Goal-based Supervisory Control}                                                                          \\
\multicolumn{1}{|l|}{\textbf{Remote Control / Operation}}                                                                   & \multicolumn{1}{c|}{Remote Driver}                                                                                                            & \multicolumn{1}{c|}{\begin{tabular}[c]{@{}c@{}}Remote Dispatcher / Operator /\\ Remote Supervisor\end{tabular}}              \\ \hline
\multicolumn{1}{|l|}{APTIV et al. \cite{aptiv2019SafetyFirstAutomated}}                                                                                          & \multicolumn{1}{c|}{Operating the vehicle}                                                                                                    & \multicolumn{1}{c|}{Indirect Teleoperation}                                                                                  \\
\multicolumn{1}{|l|}{\textbf{Teleoperation / Remote Control}}                                                               & \multicolumn{1}{c|}{Remote (Vehicle) Operator / Teleoperator}                                                                                 & \multicolumn{1}{c|}{\begin{tabular}[c]{@{}c@{}}Remote (Vehicle) Operator /\\ Teleoperator\end{tabular}}                      \\ \hline
                                                                                                                            &                                                                                                                                               &                                                                                                                              \\ \hline
\multicolumn{1}{|l|}{Nissan \cite{nissan2019SeamlessAutonomousMobility}}                                                                                                & \multicolumn{1}{c|}{-}                                                                                                                        & \multicolumn{1}{c|}{Remote Human Support}                                                                                    \\
\multicolumn{1}{|l|}{\textbf{Seamless Autonomous Mobility}}                                                                 & \multicolumn{1}{c|}{-}                                                                                                                        & \multicolumn{1}{c|}{Mobility Manager}                                                                                        \\ \hline
\multicolumn{1}{|l|}{Waymo \cite{ackerman2021WhatFullAutonomy}}                                                                                                 & \multicolumn{1}{c|}{-}                                                                                                                        & \multicolumn{1}{c|}{Fleet Response}                                                                                          \\
\multicolumn{1}{|l|}{\textbf{Fleet Response}}                                                                               & \multicolumn{1}{c|}{-}                                                                                                                        & \multicolumn{1}{c|}{\begin{tabular}[c]{@{}c@{}}Fleet Response Specialist /\\ Operator\end{tabular}}                          \\ \hline
\multicolumn{1}{|l|}{Aurora \cite{aurora_teleassist_2019}}                                                                                                & \multicolumn{1}{c|}{-}                                                                                                                        & \multicolumn{1}{c|}{Teleassist}                                                                                              \\
\multicolumn{1}{|l|}{\textbf{Teleassist}}                                                                                   & \multicolumn{1}{c|}{-}                                                                                                                        & \multicolumn{1}{c|}{Teleassist Specialist}                                                                                   \\ \hline
\multicolumn{1}{|l|}{Zoox \cite{zoox_how_2020}}                                                                                                  & \multicolumn{1}{c|}{-}                                                                                                                        & \multicolumn{1}{c|}{Remote Assistance}                                                                                       \\
\multicolumn{1}{|l|}{\textbf{TeleGuidance / TeleOps}}                                                                       & \multicolumn{1}{c|}{-}                                                                                                                        & \multicolumn{1}{c|}{\begin{tabular}[c]{@{}c@{}}Teleguidance Operators /\\ Teleoperators\end{tabular}}                        \\ \hline
\multicolumn{1}{|l|}{ARGO AI \cite{argo_ai_argo_2021}}                                                                                               & \multicolumn{1}{c|}{-}                                                                                                                        & \multicolumn{1}{c|}{Remote Guidance}                                                                                         \\
\multicolumn{1}{|l|}{\textbf{Remote Guidance}}                                                                              & \multicolumn{1}{c|}{-}                                                                                                                        & \multicolumn{1}{c|}{Remote Guidance Operator}                                                                                \\ \hline
\multicolumn{1}{|l|}{UBER \cite{aitken2019TelecommunicationsNetworkVehicles}}                                                                                                  & \multicolumn{1}{c|}{\begin{tabular}[c]{@{}c@{}}Manual Control incl. Vehicle\\ Operating Assistance Technology\end{tabular}}                   & \multicolumn{1}{c|}{Semi-Autonomous Mode}                                                                                    \\
\multicolumn{1}{|l|}{\textbf{\begin{tabular}[c]{@{}l@{}}Remote (Autonomous)\\ Vehicle Assistance System\end{tabular}}}      & \multicolumn{1}{c|}{Operator}                                                                                                                 & \multicolumn{1}{c|}{Operator}                                                                                                \\ \hline
\multicolumn{1}{|l|}{Bosch \cite{geraldy2020SystemSafeTeleoperated}}                                                                                                 & \multicolumn{1}{c|}{Drive}                                                                                                                    & \multicolumn{1}{c|}{Guide}                                                                                                   \\
\multicolumn{1}{|l|}{\textbf{Teleoperated Driving}}                                                                         & \multicolumn{1}{c|}{Remote Operator}                                                                                                          & \multicolumn{1}{c|}{Remote Operator}                                                                                         \\ \hline
\multicolumn{1}{|l|}{Baidu Apollo \cite{baidu2020BuildingSelfdrivingCar}}                                                                                          & \multicolumn{1}{c|}{Teleoperation}                                                                                                            & \multicolumn{1}{c|}{-}                                                                                                       \\
\multicolumn{1}{|l|}{\textbf{5G Remote Driving Service}}                                                                    & \multicolumn{1}{c|}{Remote Human Operator}                                                                                                  & \multicolumn{1}{c|}{-}                                                                                                       \\ \hline
\multicolumn{1}{|l|}{Valeo \cite{valeo2019ValeoDrive4URemote}}                                                                                                 & \multicolumn{1}{c|}{Teleoperation}                                                                                                            & \multicolumn{1}{c|}{-}                                                                                                       \\
\multicolumn{1}{|l|}{\textbf{Drive4U Remote}}                                                                               & \multicolumn{1}{c|}{Operator}                                                                                                                 & \multicolumn{1}{c|}{-}                                                                                                       \\ \hline
\multicolumn{1}{|l|}{Einride \cite{einride2021EinrideShowcasesOne}}                                                                                               & \multicolumn{1}{c|}{Remote Operation}                                                                                                       & \multicolumn{1}{c|}{-}                                                                                                       \\
\multicolumn{1}{|l|}{\textbf{Remote Operation}}                                                                             & \multicolumn{1}{c|}{Remote Operator}                                                                                                          & \multicolumn{1}{c|}{-}                                                                                                       \\ \hline
\multicolumn{1}{|l|}{Phantom Auto \cite{phantomauto2021HomePhantomAuto}}                                                                                          & \multicolumn{1}{c|}{Tele-Driving}                                                                                                             & \multicolumn{1}{c|}{Tele-Assistance}                                                                                         \\
\multicolumn{1}{|l|}{\textbf{Remote Operation}}                                                                             & \multicolumn{1}{c|}{Driver / Operator}                                                                                                          & \multicolumn{1}{c|}{Driver / Operator}                                                                                         \\ \hline
\multicolumn{1}{|l|}{Ottopia \cite{rosenzweig2021OttopiaUniversalTeleoperation}}                                                                                               & \multicolumn{1}{c|}{\begin{tabular}[c]{@{}c@{}}Direct Control incl. Advanced\\ Teleoperator Assistance System\end{tabular}}                   & \multicolumn{1}{c|}{\begin{tabular}[c]{@{}c@{}}Indirect Control /\\ Advanced Teleoperation\end{tabular}}                     \\
\multicolumn{1}{|l|}{\textbf{Teleoperation}}                                                                                & \multicolumn{1}{c|}{Teleoperator / Remote Operator}                                                                                           & \multicolumn{1}{c|}{Teleoperator / Remote Operator}                                                                          \\ \hline
\multicolumn{1}{|l|}{Vay \cite{vay2021}}                                                                                               & \multicolumn{1}{c|}{\begin{tabular}[c]{@{}c@{}}Teledriving / Remotely Driving\end{tabular}}                   & \multicolumn{1}{c|}{\begin{tabular}[c]{@{}c@{}}-\end{tabular}}                     \\
\multicolumn{1}{|l|}{\textbf{Teledriving}}                                                                                & \multicolumn{1}{c|}{Teledriver}                                                                                           & \multicolumn{1}{c|}{-}                                                                          \\ \hline
\multicolumn{1}{|l|}{Fernride \cite{fernride2021FernrideEnablingDriverless}}                                                                                              & \multicolumn{1}{c|}{Teleoperation}                                                                                                            & \multicolumn{1}{c|}{-}                                                                                                       \\
\multicolumn{1}{|l|}{\textbf{Teleoperation}}                                                                                & \multicolumn{1}{c|}{Remote Operator}                                                                                                          & \multicolumn{1}{c|}{-}                                                                                                       \\ \hline
                                                                                                                            & \multicolumn{1}{l}{}                                                                                                                          & \multicolumn{1}{l}{}                                                                                                         \\ \hline
\multicolumn{1}{|l|}{Sheridan \cite{sheridan1989Telerobotics}}                                                                                              & \multicolumn{1}{c|}{Teleoperation}                                                                                                            & \multicolumn{1}{c|}{Telerobotics / Supervisory Control}                                                                      \\
\multicolumn{1}{|l|}{\textbf{Teleoperation}}                                                                                & \multicolumn{1}{c|}{Human Operator}                                                                                                           & \multicolumn{1}{c|}{Human Operator}                                                                                          \\ \hline
\multicolumn{1}{|l|}{Schitz et al. \cite{schitz2020CorridorBasedSharedAutonomy}}                                                                                         & \multicolumn{1}{c|}{-}                                                                                                                        & \multicolumn{1}{c|}{Remote Operation}                                                                                        \\
\multicolumn{1}{|l|}{\textbf{Teleoperated Driving / Teleoperation}}                                                         & \multicolumn{1}{c|}{-}                                                                                                                        & \multicolumn{1}{c|}{Human Operator}                                                                                          \\ \hline
\multicolumn{1}{|l|}{Neumeier et al. \cite{neumeier2018WayAutonomousVehicles,neumeier2019MeasuringFeasibilityTeleoperated,NeumeierTeleoperation2019}}                                                                                       & \multicolumn{1}{c|}{\begin{tabular}[c]{@{}c@{}}Remote Driving / \\ Remote Control (Driving)\end{tabular}}                                      & \multicolumn{1}{c|}{-}                                                                                                       \\
\multicolumn{1}{|l|}{\textbf{Teleoperation / Teleoperated Driving}}                                                         & \multicolumn{1}{c|}{\begin{tabular}[c]{@{}c@{}}Remote Operator / \\Teleoperated Driver\end{tabular}}                                          & \multicolumn{1}{c|}{-}                                                                                                       \\ \hline
\multicolumn{1}{|l|}{Graf et al. \cite{graf2020DesignSpaceAdvanced,graf2020ImprovingPredictionAccuracy,graf2020PredictiveCorridorVirtual}}                                                                                           & \multicolumn{1}{c|}{\begin{tabular}[c]{@{}c@{}}Teleoperation / Direct Control\\ (incl. Teledriving Assistance Systems)\end{tabular}}          & \multicolumn{1}{c|}{Indirect Control}                                                                                        \\
\multicolumn{1}{|l|}{\textbf{\begin{tabular}[c]{@{}l@{}}Teleoperation / Teleoperated\\ Autonomous Vehicles\end{tabular}}}   & \multicolumn{1}{c|}{Teleoperator / (Remote) Operator}                                                                                         & \multicolumn{1}{c|}{Teleoperator / (Remote) Operator}                                                                         \\ \hline
\multicolumn{1}{|l|}{Georg et al. \cite{georg2018TeleoperatedDrivingKeya,georg2019AdaptableImmersiveReal,georg2020SensorActuatorLatency}}                                                                                          & \multicolumn{1}{c|}{Teleoperation}                                                                                                            & \multicolumn{1}{c|}{-}                                                                                                       \\
\multicolumn{1}{|l|}{\textbf{Teleoperated Driving / Teleoperation}}                                                         & \multicolumn{1}{c|}{(Remote) Operator}                                                                                                        & \multicolumn{1}{c|}{-}                                                                                                       \\ \hline
\multicolumn{1}{|l|}{Schimpe et al. \cite{schimpe2020SteerMePredictive,schimpe2021AdaptiveVideoConfiguration}}                                                                                         & \multicolumn{1}{c|}{\begin{tabular}[c]{@{}c@{}}Teleoperation / Stabilization level\\ Control (incl. Semi-Autonomous Controller)\end{tabular}} & \multicolumn{1}{c|}{Human-Machine Collaboration}                                                                             \\
\multicolumn{1}{|l|}{\textbf{Teleoperated Driving / Teleoperation}}                                                         & \multicolumn{1}{c|}{Human Operator / Teleoperator}                                                                                            & \multicolumn{1}{c|}{Human Operator / Teleoperator}                                                                           \\ \hline
\multicolumn{1}{|l|}{HF-IRADS \cite{hf-irads2020HumanFactorsChallenges}}                                                                                              & \multicolumn{1}{c|}{Remote Management}                                                                                                        & \multicolumn{1}{c|}{Remote Control}                                                                                          \\
\multicolumn{1}{|l|}{\textbf{Remote Support and Control}}                                                                   & \multicolumn{1}{c|}{Remote Human / Operator}                                                                                                  & \multicolumn{1}{c|}{Remote Human / Operator}                                                                                 \\ \hline
\multicolumn{1}{|l|}{Mutzenich et al. \cite{mutzenich2021UpdatingOurUnderstanding}}                                                                                      & \multicolumn{1}{c|}{Remote Control / Teleoperation}                                                                                           & \multicolumn{1}{c|}{\begin{tabular}[c]{@{}c@{}}Remote Management / \\Goal-based Supervision\end{tabular}}                    \\
\multicolumn{1}{|l|}{\textbf{Remote Operation}}                                                                             & \multicolumn{1}{c|}{Remote Operator / Driver}                                                                                                 & \multicolumn{1}{c|}{Remote Operator / Controller}                                                                            \\ \hline
\multicolumn{1}{|l|}{Fennel et al. \cite{fennel2021HapticGuidedPathGeneration}}                                                                                         & \multicolumn{1}{c|}{-}                                                                                                                        & \multicolumn{1}{c|}{Shared Autonomy}                                                                                         \\
\multicolumn{1}{|l|}{\textbf{Teleoperation}}                                                                                & \multicolumn{1}{c|}{-}                                                                                                                        & \multicolumn{1}{c|}{Human Operator}                                                                                          \\ \hline
\multicolumn{1}{|l|}{Kettwich et al. \cite{kettwich2021TeleoperationHighlyAutomated}}                                                                                       & \multicolumn{1}{c|}{Direct Approach}                                                                                                          & \multicolumn{1}{c|}{Indirect Approach}                                                                                       \\
\multicolumn{1}{|l|}{\textbf{\begin{tabular}[c]{@{}l@{}}Teleoperation / Remote-Operation /\\ Remote-Control\end{tabular}}}         & \multicolumn{1}{c|}{Remote-Operator / Teleoperator}                                                                                           & \multicolumn{1}{c|}{Remote-Operator / Teleoperator}                                                                          \\ \hline
\multicolumn{1}{|l|}{Hofbauer et al. \cite{hofbauer2020TELECARLAOpenSource}}                                                                                       & \multicolumn{1}{c|}{\begin{tabular}[c]{@{}c@{}}Teleoperated Driving / Teleoperation /\\ Teledriving\end{tabular}}                             & \multicolumn{1}{c|}{-}                                                                                                       \\
\multicolumn{1}{|l|}{\textbf{\begin{tabular}[c]{@{}l@{}}Teleoperated Driving / Teleoperation /\\ Teledriving\end{tabular}}} & \multicolumn{1}{c|}{Teleoperator / Human Operator}                                                                                            & \multicolumn{1}{c|}{-}                                                                                                       \\ \hline
                                                                                                                            & \multicolumn{1}{l}{}                                                                                                                          & \multicolumn{1}{l}{}                                                                                                         \\ \hline
\multicolumn{1}{|l|}{German Government \cite{bundesregierung_entwurf_2021}}                                                                                     & \multicolumn{1}{c|}{-}                                                                                                                        & \multicolumn{1}{c|}{\begin{tabular}[c]{@{}c@{}}Authorization of driving maneuvers,\\ provision of alternatives\end{tabular}} \\
\multicolumn{1}{|l|}{-}                                                                                                     & \multicolumn{1}{c|}{-}                                                                                                                        & \multicolumn{1}{c|}{Technical Supervisor}                                                                                    \\ \hline
\multicolumn{1}{|l|}{UK Government \cite{ukgovernment2019CodePracticeAutomated}}                                                                                         & \multicolumn{1}{c|}{Remote-Control Function}                                                                                                  & \multicolumn{1}{c|}{-}                                                                                                       \\
\multicolumn{1}{|l|}{\textbf{Remote-controlled Operation}}                                                                  & \multicolumn{1}{c|}{(Remote) Safety Driver / Operator}                                                                                        & \multicolumn{1}{c|}{-}                                                                                                       \\ \hline
\multicolumn{1}{|l|}{State of California \cite{departmentofmotorvehiclescaliforniausa2021ArticleTestingAutonomous}}                                                                                   & \multicolumn{1}{c|}{Performance of the DDT}                                                                                                   & \multicolumn{1}{c|}{-}                                                                                                       \\
\multicolumn{1}{|l|}{\textbf{-}}                                                                                            & \multicolumn{1}{c|}{Remote Operator}                                                                                                          & \multicolumn{1}{c|}{-}                                                                                                       \\ \hline
\multicolumn{1}{|l|}{State of  Arizona \cite{stateofarizona2015ExecutiveOrder201509SelfDrivingVehicle}}                                                                                     & \multicolumn{1}{c|}{Direct a vehicle's movement}                                                                                              & \multicolumn{1}{c|}{Provide direction}                                                                                       \\
\multicolumn{1}{|l|}{\textbf{-}}                                                                                            & \multicolumn{1}{c|}{Operator}                                                                                                                 & \multicolumn{1}{c|}{Operator}                                                                                                \\ \hline
\end{tabular}%
		}
	\newline
	\caption{Overview of RHIS terminologies from standards and technical reports, corporates, scientific publications and legislative bills}
	\label{tab:survey}
\end{table}

\subsection{Standards and Technical Reports}
The SAE standard J3016 \cite{sae_j3016c_2021} has been defining taxonomies on DAS since 2014, leading to the levels 0-5 which are the de-facto standard in systemizing capabilites of DAS. With the latest revision J3016\textsubscript{202104}, the standard has also defined the terms \emph{remote assistance} and \emph{remote driving}. While no general term is introduced, \cite{2021SAELevelsDriving} lists those under the category \emph{remote support functions}.

Both \emph{remote driving} and \emph{remote assistance} are closely linked to the \emph{dynamic driving task} (DDT) specified by the standard, which includes the following tasks:

\begin{itemize}
	\item Object and event detection and response (OEDR),
	\item Tactical maneuver planning,
	\item Lateral and longitudinal motion control,
	\item Enhancing conspicuity via lighting [...] etc.
\end{itemize}

Level 3-5 demand that the ADS is performing the whole DDT, while only level 4-5 need to be able to additionally perform \emph{DDT fallbacks}, which are necessary in case of system failures or the exit of the operational design domain (ODD). These fallbacks lead to the \emph{minimal risk condition}, which is "a stable, stopped condition [...] in order to reduce the risk of a crash [...]" on their own. In contrast, level 3 ADS can rely on a human \emph{fallback-ready user}.

\emph{Remote assistance} is defined as "event-driven provision, by a remotely located human, of information [...] in order to facilitate trip continuation when the ADS encounters a situation it cannot manage." In such situations, the ADS performs the DDT at all times and no real-time requirements for the human support exist. Examples are provision of pathways or revised goals, and classification of objects\footnote{While classification of objects might be a task performed by a remote assistant, they cannot provide this classification to the ADS, since it is part of the DDT, which needs to be performed by the ADS. Since it is unclear how to classify RHIS which provide information about the environment, we will assign them to level 4-5.}.
\emph{Remote assistance} is only available at levels 4 and 5. It includes supporting the DDT as well as the DDT fallback. 

\emph{Remote driving} is defined as "real-time performance of part or all of the DDT and/or DDT fallback [...] by a \emph{remote driver}." The standard recommends against using the term \emph{teleoperation} instead of \emph{remote driving}, since it is "not defined consistently in the literature". \emph{Remote driving} is only available at levels 1 and 2. A special case exists for level 3, where a \emph{(remote) fallback-ready user} is necessary. When performing the fallback remotely, such a user immediately becomes a \emph{remote driver}. Since the \emph{(remote) fallback-ready user} only performs a special form of (passive) monitoring, we will not further include level 3 within the survey, but include the role within the monitoring section in the taxonomy found in Section \ref{sec:taxonomy}.\\

Cummings et al. \cite{cummings2020ConceptsOperationsAutonomous} identify in their \emph{CSCRS Concepts of Operations for Autonomous Vehicle Dispatch Operations} report \emph{remote control} as a basic function necessary for ADS and use the term \emph{remote operation} interchangeably. They classify existing approaches into two categories: \emph{Teleoperation} and \emph{Goal-based supervisory control}.
In regards to \emph{teleoperation}, they have the understanding that a \emph{remote driver} takes over full control. However, they see latencies (of the human and the network) as critical, which is why they do not consider level 3 use to be reasonable. For level 1-2, they mostly consider slow driving tasks up to ~15 km/h. 
\emph{Goal-based supervisory control} is how they name services in which a \emph{remote dispatcher}, also called \emph{operator} or \emph{remote supervisor}, provides the ADS with goals, which the ADS executes on its own. They see disadvantages in situations where a sensor of the vehicle is degraded, which might affect the performance of the ADS significantly.\\

In the \emph{Safety First for Automated Driving} \cite{aptiv2019SafetyFirstAutomated} report, an international consortium of eleven companies provides an overview of "safety by design and verification and validation" methods of level 3-4 ADS. While they put no special emphasis on RHIS, \emph{teleoperation}, also called \emph{remote control}, by \emph{remote (vehicle) operators} or \emph{teleoperators} is included in their analysis. They differentiate between \emph{operating the vehicle} and \emph{indirect teleoperation}, without going into further detail. 

\subsection{Corporate approaches}

In this section we refer to approaches of companies around the world, be it on a concept level or already implemented and running. Most of these approaches have not been published in scientific papers, but as part of PR campaigns, safety reports or in the form of patents. However, since we are primarily looking at the systematic level in this survey, this does not pose a problem, as details on technical implementations are secondary. While several companies have dissolved or been acquired, we always refer to the original source.\\

The \emph{Nissan Seamless Autonomous Mobility} \cite{nissan2019SeamlessAutonomousMobility} technology provides \emph{remote human support} for ADS in cases when "human judgment is required for the appropriate course". After the vehicle has brought itself to a safe stop, vehicle sensor data is streamed to a \emph{mobility manager}, who remotely provides the vehicles with a path or lane to follow.

\emph{Bosch} \cite{geraldy2020SystemSafeTeleoperated} proposes \emph{teleoperated driving}, where an \emph{external operator} can either \emph{guide} the ADS or {drive} the vehicle.

\emph{Valeo} demonstrated the \emph{Drive4U Remote} \cite{valeo2019ValeoDrive4URemote}
service, where an \emph{operator} can \emph{teleoperate} the vehicle, i.e. to control a vehicle directly. 
 
The Waymo level 4 ADS, also called \emph{Waymo Driver}, is currently publicly available in Phoenix, Arizona, USA \cite{ackerman2021WhatFullAutonomy}. In situations where human input is necessary, it can call on \emph{fleet response}. The \emph{fleet response specialists}, also called \emph{fleet response operators} \cite{waymo2021SoftwareEngineerFleet}, will then support the vehicle with strategical input, which is not part of the DDT, such as waypoints.

The \emph{Aurora Teleassist} \cite{aurora_teleassist_2019} system utilizes the input of so called \emph{teleassist specialists}. They will have access to raw and abstracted sensor data, when the ADS, also called \emph{Aurora Driver}, is unsure on how to continue. Based on this information, they will be able to support the ADS on the levels "mapping, perception, route planning, and trajectory planning". The ADS will then incorporate this information to continue the trip.

The \emph{Zoox TeleGuidance} \cite{zoox_how_2020} system, also called \emph{TeleOps}, gets activated as soon as the ADS encounters rare, unusual or complex scenarios it cannot handle on its own. \emph{Teleguidance operators}, also called \emph{teleoperators}, then provide \emph{remote assistance}. They have access to raw camera data and abstracted sensor data and interact with the system via a touchscreen. Examples for their guidance are provision of suggested paths and labeling objects.

The \emph{Voyage Telesisst} \cite{cameron_introducing_2020} is a multi-level system for \emph{remote assistance}. When the ADS is unsure on how to proceed, \emph{discrete decision making} can be activated for \emph{remote assistance}. On this level, \emph{ROp} provide high-level inputs, which can range from simple commands, such as "Proceed", to provision of a path to follow.
In cases where this does not suffice, \emph{remote driving} can be activated. In this case, a \emph{ROp} is directly controlling the car. To avoid collisions and dangerous situations, an additional \emph{remote drive assist} monitors the state of the vehicle and avoids collisions by stopping it or limiting the maximum speed. The \emph{ROp} are located within a so-called "telessist pod", which is a workstation that consists of components for steering, throttle, and brake as well as a touchscreen. Operators receive raw camera data and an abstracted environment model and have additional access to the state of the vehicle and the behavior planner outputs.

The \emph{ARGO AI Remote Guidance} system \cite{argo_ai_argo_2021} is activated when the ADS requires additional guidance, in which case a \emph{remote guidance operator} can provide "authorization to the [ADS] to perform specific driving tasks it has recommended". Thus, the \emph{remote guidance operator} does not provide additional input to the system, but authorizes an appropriate action suggested by the ADS.

The \emph {UBER remote vehicle assistance system}  \cite{aitken2019TelecommunicationsNetworkVehicles}, also called \emph{autonomous vehicle assistance system} or \emph{remote autonomous vehicle assistance} \cite{chen2020ContextBasedRemoteAutonomous}, allows an ADS to request \emph{remote assistance} if it encounters problems. A \emph{human operator} can \emph{manually control} the ADS, while being supported by a \emph{vehicle operating assistance technology} that can provide collision mitigation. The ADS can also operate in a \emph{semi-autonomous mode}, in which the \emph{human operator} can provide \emph{task management data} \cite{chen2020ContextBasedRemoteAutonomous}. As examples, object classification and the provision of waypoints to follow are listed. 

\emph{Baidu Apollo} integrates a \emph{5G remote driving service} \cite{baidu2020BuildingSelfdrivingCar}, which they also refer to as \emph{5G-enabled teleoperation}. \emph{Remote human operators} can directly control the vehicles if necessary.

\emph{Einride} \cite{einride2021EinrideShowcasesOne} showcases a \emph{remote operation} service, in which \emph{ROp} can take control over a vehicle from a \emph{remote drive station}.

\subsubsection{Dedicated RHIS Companies}

\emph{Phantom Auto} \cite{phantomauto2021HomePhantomAuto} provides a service for \emph{remote operation} of a wide variety of vehicles. \emph{(Remote) drivers/operators} can interact via \emph{tele-driving}, which enables "real-time, direct remote control" and \emph{tele-assistance}, where they "remotely guide and command automated functions". Drivers typically have HD video data and a steering wheel as the interface.

\emph{Ottopia} \cite{rosenzweig2021OttopiaUniversalTeleoperation} provides services for \emph{teleoperation}, by which they define \emph{direct} and \emph{indirect control}. Their \emph{advanced teleoperation (ATO)} suite provides \emph{indirect methods} such as the proposition of possible paths from which to choose from or the provision of a reference-path to follow. \emph{Direct control} on the other hand is supported by their \emph{advanced teleoperator assistance systems (ATAS)}, which is a collision-avoidance system while a \emph{teleoperator}, also called \emph{ROp}, is directly controlling the vehicle.

\emph{Fernride} \cite{fernride2021FernrideEnablingDriverless} provides a \emph{teleoperation} solution where \emph{remote operators} directly control vehicles.

\emph{Vay} \cite{vay2021} uses a similar concept, where \emph{teledrivers} remotely drive, also called \emph{teledrive}, remote vehicles.

\subsection{Research approaches}

Already in 1989 there was confusion about the term \emph{teleoperation}. Sheridan \cite{sheridan1989Telerobotics} describes \emph{teleoperation} very general as "extension of a person's sensing and manipulation capability to a remote location", but also admits that "\emph{teleoperation} refers most commonly to direct and continuous human control [...], but can also be used generally to encompass \emph{telerobotics} as well." \emph{Telerobotics} then is described as "[...] a form of teleoperation in which a human operator acts as a supervisor, intermittently communicating to a computer information about goals, constraints, plans, contingencies, assumptions, suggestions and orders relative to a limited task, getting back information about accomplishments, difficulties, concerns, and, as requested, raw sensory data while the subordinate telerobot executes the task based on information received from the human operator plus its own artificial sensing and intelligence." As a synonym, \emph{supervisory control} is utilized as a broader term for "any semi-autonomous system". It becomes obvious that already in 1989 the concepts of direct and indirect RHIS were well known. While most of the technologies of DAS stem from the field of robotics, this paper puts a focus on current terminologies and therefore examines recent publications in relation to DAS. \\

Beginning in 2018, Georg et al. primarily utilize the term \emph{teleoperated driving}, also sometimes referred to as \emph{teleoperation}, together with \emph{(remote) operator}, also referred to as \emph{human operator} in multiple publications \cite{georg2018TeleoperatedDrivingKeya} \cite{georg2019AdaptableImmersiveReal} \cite{georg2020SensorActuatorLatency} consistently to describe direct control approaches. In the following we list multiple authors from the same research group. Schimpe et al. \cite{schimpe2020SteerMePredictive,schimpe2021AdaptiveVideoConfiguration} follow the terms, while also utilizing the term \emph{teleoperator}. In \cite{schimpe2020SteerMePredictive} they also distinguish between direct \emph{stabilization level control} and indirect \emph{human-machine collaboration}. They also introduce a \emph{semi-autonomous controller} "to correct the steering input given by the teleoperator if the vehicle is at risk of hitting an obstacle." Saparia et al. \cite{saparia2021ActiveSafetySystem} and Hoffmann and Diermeyer \cite{hoffmann2021SystemsTheoreticSafetyAssessment} also follow these terms. \\

In 2019, Neumeier et al. \cite{NeumeierTeleoperation2019} performed a user study on latency-effects for direct \emph{teleoperation}. They use the terms \emph{teleoperated driving},  \emph{remote driving} and \emph{remote control} interchangeably and also refer to the broader term \emph{remote operation}. \emph{ROp}, also called \emph{teleoperated drivers} are introduced as "remote operating fallback authority". In \cite{neumeier2019MeasuringFeasibilityTeleoperated}, Neumeier et al. also introduce the term \emph{remote control driving}. \\

In 2020, Schitz et al. \cite{schitz2020CorridorBasedSharedAutonomy} demonstrate an approach where a remote \emph{human operator} provides an ADS with a corridor instead of waypoints, in which the ADS computes a collison-free path by itself. The authors use the terms \emph{teleoperated driving}, \emph{teleoperation} and \emph{remote operation} interchangeably.

Hofbauer et al. \cite{hofbauer2020TELECARLAOpenSource} provide a simulation-based framework for direct \emph{teleoperation}, which they also refer to as \emph{teleoperated driving} or \emph{teledriving} by a \emph{teleoperator}, also called \emph{human operator}.

Graf et al. \cite{graf2020PredictiveCorridorVirtual} utilize the terms \emph{teleoperated autonomous vehicles}, \emph{teleoperation} and \emph{teleoperator} respectively \emph{(remote) operator} consistently for direct control methods. In \cite{graf2020ImprovingPredictionAccuracy} they show progress on \emph{teledriving assistance systems} that support direct control methods to improve situational awareness. Only in \cite{graf2020DesignSpaceAdvanced} they broaden the term \emph{teleoperation} to include both \emph{direct control} and \emph{indirect control} as \emph{driving modes}. Examples for \emph{indirect control} include maneuver selection or provision of a trajectory. 

A position paper from HF-IRADS \cite{hf-irads2020HumanFactorsChallenges} defines three major categories for \emph{Remote Support and Control} by \emph{remote humans}, also called \emph{ROp}:
\begin{itemize}
	\item \emph{Remote Assistance} for passenger-related services and breakdown-assistance  
	\item \emph{Remote Management} for indirect assistance of the ADS
	\item \emph{Remote Control} for full remote control over the DDT
\end{itemize}

Mutzenich et al. \cite{mutzenich2021UpdatingOurUnderstanding} follow the categories provided by \cite{hf-irads2020HumanFactorsChallenges}, while using the term \emph{goal-based supervision} interchangeably with \emph{remote management} by a \emph{remote controller}. The same holds for \emph{teleoperation} and \emph{remote control} by a \emph{remote driver}. They propose a revised version of the J3016\textsubscript{201806} \cite{saeinternational2018J3016BTaxonomyDefinitions} from 2018 to include \emph{remote interventions} at levels 3, 4 and 5. While their proposal, published on February 19, 2021, was not adopted identically, it is worth noting that the on May 3, 2021 updated SAE standard \cite{sae_j3016c_2021} does include \emph{remote support functions}.

Fennel et al. \cite{fennel2021HapticGuidedPathGeneration} use the term \emph{teleoperation} in a broad fashion but present an indirect \emph{shared autonomy} approach, where a \emph{human operator} provides a reference path.

Kettwich et al. \cite{kettwich2021TeleoperationHighlyAutomated} use the terms \emph{teleoperation} and \emph{remote-operation / control} interchangeably. They separate \emph{direct} and \emph{indirect approaches}, describing the latter as \emph{human interventions}. They primarily use the term \emph{remote-operator}, while occasionally referring to them as \emph{remote human} or \emph{teleoperator}.

\subsection{Legislative Frameworks}
In May 2021, The German government has laid the legal conditions for level 4-5 driving in regular operation \cite{bundesregierung_entwurf_2021}. In cases where the vehicles cannot continue on their own, a \emph{technical supervisor} can assist remotely. Their main task is to either deactivate the vehicle or authorize possible driving maneuvers provided by the ADS. Additionally, the \emph{technical supervisor} can provide alternative driving maneuvers for the ADS.

In the \emph{Autonomous Vehicle Policy Framework: Selected National and Jurisdictional Policy Efforts to Guide Safe AV Development} report \cite{worldeconomicforum2020AutonomousVehiclePolicy} the situation in Israel, Singapore, UK, Australia and the US states California and Arizona is summarized. Only the UK and both US states have regulation in regards to RHIS. 

In the UK \emph{Code of Practice: Automated vehicle trialling}, RHIS is summarized as \emph{remote-controlled operation} \cite{ukgovernment2019CodePracticeAutomated}. A \emph{safety driver}, who can also be located remotely, or a \emph{(remote) safety operators} who is "ready and able to override the vehicle" if necessary is required. \emph{Remote-controlled operation}, which is a general term for \emph{remote-control functions}, is a direct approach, described as "two-way, real-time communication".

In the US state California, performance of the DDT can be done by a \emph{ROp}, who can also command the ADS to get into a minimal risk condition \cite{departmentofmotorvehiclescaliforniausa2021ArticleTestingAutonomous}. In Arizona, an \emph{operator} can monitor and "provide direction remotely" respectively "direct a vehicle's movement" \cite{stateofarizona2015ExecutiveOrder201509SelfDrivingVehicle}. While the executive order was updated in 2018 \cite{stateofarizona2018ExecutiveOrder201804}, no further information on RHIS is available. As the wording is very vague, we assume that both direct and indirect approaches can be meant.

\section{Taxonomy on Remote Human Input Systems}
\label{sec:taxonomy}

We propose the terminology \emph{remote human input systems} in the context of \emph{driving automation systems} for all situations, such as corner cases, where a remote human, called \emph{remote operator} in this taxonomy, is needed to continue the DDT, if necessary. The determination of when RHIS becomes necessary is outside the scope of this work. While terms as \emph{remote driving}, \emph{remote guidance}, \emph{remote assistance}, \emph{remote support}, \emph{teleoperation} and more can be found in the literature, they are not well suited as general terms, since they only include very specific types of input. The formulation \emph{remote human input systems} is agnostic to the input form and suited for all cases.

\begin{figure}[h]
	\centering
	\includegraphics[width=1\textwidth]{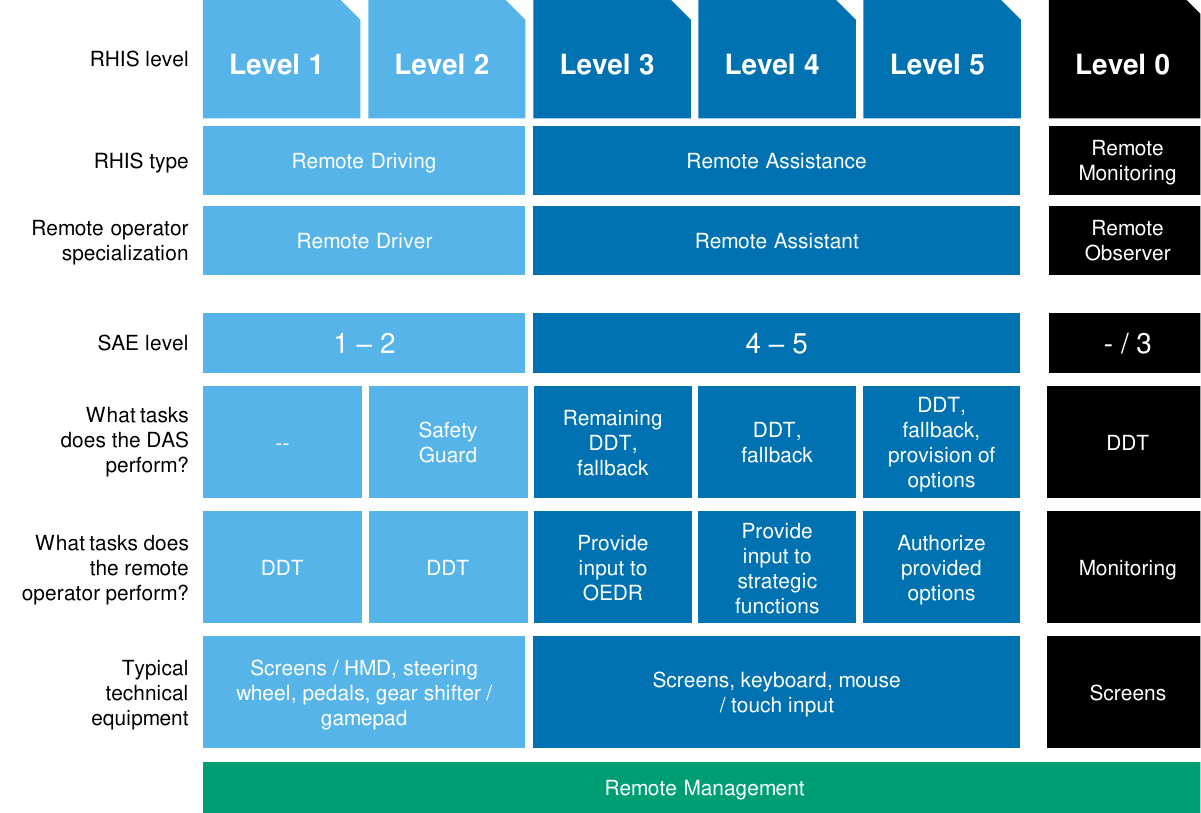}
	\caption{Taxonomy on \emph{remote human input systems} (RHIS), with RHIS levels 1 to 5 and 0. \emph{Remote operators} (ROp) provide different types of input to the \emph{driving automation system} (DAS). RHIS level 1 and 2 are of type \emph{remote driving} where a \emph{remote driver} performs the \emph{dynamic driving task} (DDT). In RHIS level 3 to 5 the ROp provides assistive input, like authorizing provided options. Level 0 is a non-interactive state only used for remote monitoring, such as diagnosing system states.}
	\label{fig:taxonomy}
\end{figure}

Centered around our proposed RHIS levels, Fig. \ref{fig:taxonomy} provides both a classification of the different levels by RHIS type, their relationship to the SAE J3016\textsubscript{202104} levels \cite{sae_j3016c_2021}, as well as naming conventions and descriptions and thus forms the taxonomy on RHIS. In the following we discuss the dimensions of this taxonomy.

\subsection{RHIS Levels}

RHIS levels are designed to assess the sophistication of a RHIS. Going from left to right in Figure \ref{fig:taxonomy}, the DAS takes on more and more tasks. RHIS level 1 to 5 describe remote systems where the ROp has the possibility to actively control the vehicle or interact with the ADS. Levels rise according to the intelligence of the ADS needed. DS systems with no ADS capability require the complete DDT to be executed by the ROp and thus are located at RHIS levels 1 and 2. Levels 3 and 4 describe higher-level input from a ROp. If only provided options need to be chosen by the ROp, such systems are of RHIS level 5. RHIS level 0 has a special meaning and is only used for monitoring the ADS remotely.

\subsection{RHIS Types}

There is no distinct remote operation technique but several operation modes. The operation mode with no interaction from the ROp to the ADS is called \emph{remote monitoring}, only present at RHIS level 0. Here, the ROp receives data from the ADS and has no possibility to intervene. The received data can be anything from just diagnostic information on the vehicles state, localization information, abstracted environmental data to complete sensor data like camera images. These information are used by the ROp to check and surveil. Remote monitoring is the basis for all other RHIS types which, in addition, allow for intervention by the ROp. The readiness to be available as a "remote fallback-ready user" also falls into RHIS level 0 and corresponds to SAE level 3.\\
The levels 1 and 2 fall into the category \emph{remote driving}, where a ROp directly controls the vehicle in the form of steering and throttle/brake commands. In level 2, the DAS can additionally control the input, avoiding collisions. In its true meaning, the term \emph{teleoperation} refers to this category.\\
Levels 3 to 5 are classified as \emph{remote assistance}. In level 3, the ROp can support the OEDR task of an ADS, e.g. by classifying objects or interpreting situations, such as hand gestures. In level 4, the ROp suggests abstracted driving commands, like goals or waypoints, which the ADS needs to interpret and execute. In level 5, the ADS provides available options, which the ROp authorizes.

\subsection{Vehicle Types}


The SAE J3016\textsubscript{202104} \cite{sae_j3016c_2021} standard  defines the \emph{(motor) vehicle} as a non-human user, comprising \emph{conventional vehicles}, \emph{(ADS-equipped) dual-mode vehicles} as well as \emph{ADS-dedicated vehicles (ADS-DV)}. Whereas \emph{conventional vehicles} are always operated by an \emph{in-vehicle driver} and \emph{ADS-dedicated vehicles} are designed for driverless operation for all trips, \emph{(ADS-equipped) dual-mode vehicles} allow for both "driverless operation under routine/normal operating conditions" and for "operation by an in-vehicle driver". This dual mode eventually allows for the completion of all kinds of trips, even after the ADS exits its ODD. Furthermore, level 5 vehicles which offer an interface for manual operation are classified as \emph{dual-mode vehicles}. The RHIS levels are vehicle agnostic, thus all types are supported by the taxonomy.

\subsection{Technical Equipment}

The required technical equipment necessary for RHIS may vary strongly, depending on the user's specific task. For \emph{remote monitoring}, a basic setup consisting of one or multiple screens displaying vehicle information or camera streams for monitoring purposes can be sufficient. 
In contrast, \emph{remote driving} also requires additional hardware for sending inputs to the vehicle. Therefore, a simple keyboard and mouse input is not sufficient, which is why most often a steering wheel, pedals and a gear shifter are installed at the remote side as for example shown in \cite{NeumeierTeleoperation2019}. For \emph{remote assistance}, where high-level inputs such as waypoints or paths are provided by a ROp, classical input components are sufficient.\\

In addition to a basic setup consisting of multiple computer screens, remote driving using a head-mounted display (HMD) has also been investigated. Krückel et. al \cite{Krueckel2015} use the HMD to display the current field of view from the perspective of the remote robot, which is extracted from a spherical camera. This allows the operator to look around freely. Kot and Novak \cite{Kot2014} use a HMD to render images from a stereovision camera, allowing for a better depth estimation of the operator. In addition, they simultaneously display complementing information within the operator's field of view, for example an image from a rear camera and additional status icons.

Furthermore, HMD can not only be used to show the real camera's video streams, but also complementary data such as lidar pointclouds or the highlighted boundaries of the driving lane. A recent user study has shown that using a HMD does not automatically increase remote driving performance compared to using conventional screens, however, additional information can be visualized in an intuitive way \cite{georg2018TeleoperatedDrivingKeya}. 

\subsection{Human Actors}
The SAE J3016\textsubscript{202104} \cite{sae_j3016c_2021} standard defines five categories of human actors: \emph{(Remote) driver}, \emph{passenger}, \emph{DDT fallback-ready user}, \emph{driverless operation dispatcher}, and \emph{remote assistant}. For RHIS the focus lies on \emph{remote drivers} and \emph{remote assistance}. We have adopted these terms in the RHIS taxonomy, since they are fitting in regards to the included functionalities in our categories. A \emph{(remote) driver} "performs in real time part or all of the DDT and/or DDT fallback for a particular vehicle", meaning he manually controls throttle, brake, steering and gear selection. A \emph{(remote)driver} becomes a \emph{DDT fallback-ready user} during the engagement of a level 3 ADS. He has to be able to respond to \emph{requests to intervene} by taking over control of the vehicle.\\

In this taxonomy we extend the human users of a DAS by \emph{remote operator (ROp)} as a general term for both \emph{remote driver},  \emph{remote assistant} and \emph{remote observer}. Depending on the RHIS level the ROp is a \emph{remote driver} when performing the complete DDT. The \emph{remote assistant} improves upon the SAE understanding of a \emph{remote assistant} who cannot possibly assist with the DDT. In the RHIS taxonomy, a \emph{remote assistant} can additionally assist with the OEDR task, as seen in many examples across the literature. The \emph{remote observer} just monitors and diagnoses the ADS and is located at RHIS level 0. Typically, ROp are trained individuals.\\

Our taxonomy focuses strongly on DAS-centric RHIS in relation to the DDT. Nevertheless, as also described in \cite{sae_j3016c_2021,hf-irads2020HumanFactorsChallenges}, there are further remote roles for dispatching or interaction with \emph{passengers}. These roles are broadly classified as \emph{remote management}.

\section{Conclusion and Future Work}

The first part (Sec. \ref{sec:survey}) of this article presents a survey on \emph{remote human input systems} (RHIS) for driving automation systems (DAS). Different terminologies exist in the literature, making it hard to compare these approaches. As part of this survey we perform an especially meticulous analysis of the most recent SAE J3016\textsubscript{202104} standard, which as we found has an insufficient definition of \emph{remote assistance}, which offers no possibility to classify many approaches presented in the survey.\\

The second part (Sec. \ref{sec:taxonomy}) introduces a taxonomy on RHIS. The taxonomy of RHIS provided allows to distinguish between the different RHIS techniques and makes comparing approaches possible. We have put a special focus on combining approaches from academia, industry and legislation. Our taxonomy introduces RHIS levels that can also serve as input for further revisions of the SAE standard.\\

Questions arise when considering the delimitation of the DAS from the ROp on a functional and also temporal perspective. Also, there is not much research on the relation between the type of event, which lead to a RHIS activation, and the RHIS classification. Such questions will be addressed in future work.

\section*{Abbreviations}
\begin{itemize}
    \item[] \makebox[2cm][l]{\textbf{ADS}} Automated Driving System
    \item[] \makebox[2cm][l]{\textbf{ADS-DV}} Automated Driving System-Dedicated Vehicle
    \item[] \makebox[2cm][l]{\textbf{DAS}} Driver Automation System
    \item[] \makebox[2cm][l]{\textbf{DDT}} Dynamic Driving Task
    \item[] \makebox[2cm][l]{\textbf{DS}} Driver Support
    \item[] \makebox[2cm][l]{\textbf{ODD}} Operational Design Domain
    \item[] \makebox[2cm][l]{\textbf{OEDR}} Object and event detection and response
    \item[] \makebox[2cm][l]{\textbf{HMD}} Head-mounted display
    \item[] \makebox[2cm][l]{\textbf{RHIS}} Remote Human Input System
    \item[] \makebox[2cm][l]{\textbf{ROp}} Remote Operator
\end{itemize}

\paragraph{Acknowledgment.}
This work results from the KIGLIS project supported by the German Federal Ministry of Education and Research (BMBF), grant number 16KIS1231.

\bibliographystyle{bibtex/spmpsci.bst} 
\bibliography{references}
\end{document}